\documentclass[11pt,a4paper]{article}
\usepackage[hyperref]{acl2021}
\usepackage{times}
\usepackage{latexsym}

\usepackage{microtype}
\usepackage{color}
\usepackage {amsmath}
\usepackage{graphicx}  % DO NOT CHANGE THIS
\usepackage{natbib}  % DO NOT CHANGE THIS AND DO NOT ADD ANY OPTIONS TO IT
\usepackage{caption} % DO NOT CHANGE THIS AND DO NOT ADD ANY OPTIONS TO IT
\usepackage{amsfonts}
\usepackage{amssymb}
\usepackage{multirow}
\usepackage{adjustbox}
\usepackage[english]{babel} % American English
\usepackage{arydshln}
\usepackage[ruled,linesnumbered]{algorithm2e}
\usepackage{bbding}
\usepackage{CJKutf8}  
\usepackage{dsfont}

\aclfinalcopy % Uncomment this line for the final submission
\def\aclpaperid{3372} %  Enter the acl Paper ID here

\title{GL-GIN: Fast and Accurate Non-Autoregressive Model for Joint Multiple Intent Detection and Slot Filling
}

\author{Libo Qin, 
	Fuxuan Wei,
	Tianbao Xie,	
	Xiao Xu,
	Wanxiang Che\thanks{\ \  Corresponding author.},
	Ting Liu\\
  Research Center for Social Computing and Information Retrieval \\
  Harbin Institute of Technology, China \\
  \texttt{\{lbqin,fuxuanwei,tianbaoxie,xxu,car,tliu\}@ir.hit.edu.cn}\\
}

\date{}

\begin{document}
\maketitle
\begin{abstract}
Multi-intent SLU can handle multiple intents in an utterance, which has attracted increasing attention.
However, the state-of-the-art joint models heavily rely on autoregressive approaches, resulting in two issues: \textit{slow inference speed} and \textit{information leakage}. 
In this paper, we explore a non-autoregressive model for joint multiple intent detection and slot filling, achieving more fast and accurate.
Specifically, we propose a \textbf{G}lobal-\textbf{L}ocally \textbf{G}raph \textbf{I}nteraction \textbf{N}etwork (GL-GIN) where a local slot-aware graph interaction layer is proposed to model slot dependency for alleviating uncoordinated slots problem while a global intent-slot graph interaction layer is introduced to model the interaction between multiple intents and all slots in the utterance. 
Experimental results on two public datasets show that our framework achieves state-of-the-art performance while being 11.5 times faster.
  \end{abstract}

\section{Introduction}
\label{Introduction}
Spoken Language Understanding (SLU)~\citep{young2013pomdp} is a critical component in spoken dialog systems, which aims to understand user's queries.
It typically includes two sub-tasks: intent detection and slot filling~\citep{tur2011spoken}.

Since intents and slots are closely tied, dominant single-intent SLU systems in the literature~\citep{goo-etal-2018-slot, li-etal-2018-self, liu-etal-2019-cm, e-etal-2019-novel, qin-etal-2019-stack,teng2021injecting,qin2021cointeractive,qin2021survey} adopt joint models to consider the correlation between the two tasks, which have obtained remarkable success.

Multi-intent SLU means that the system can handle an utterance containing multiple intents, which is shown to be more practical in the real-world scenario, attracting increasing attention.
\begin{figure}[t]
	\centering
	\includegraphics[width=0.44\textwidth]{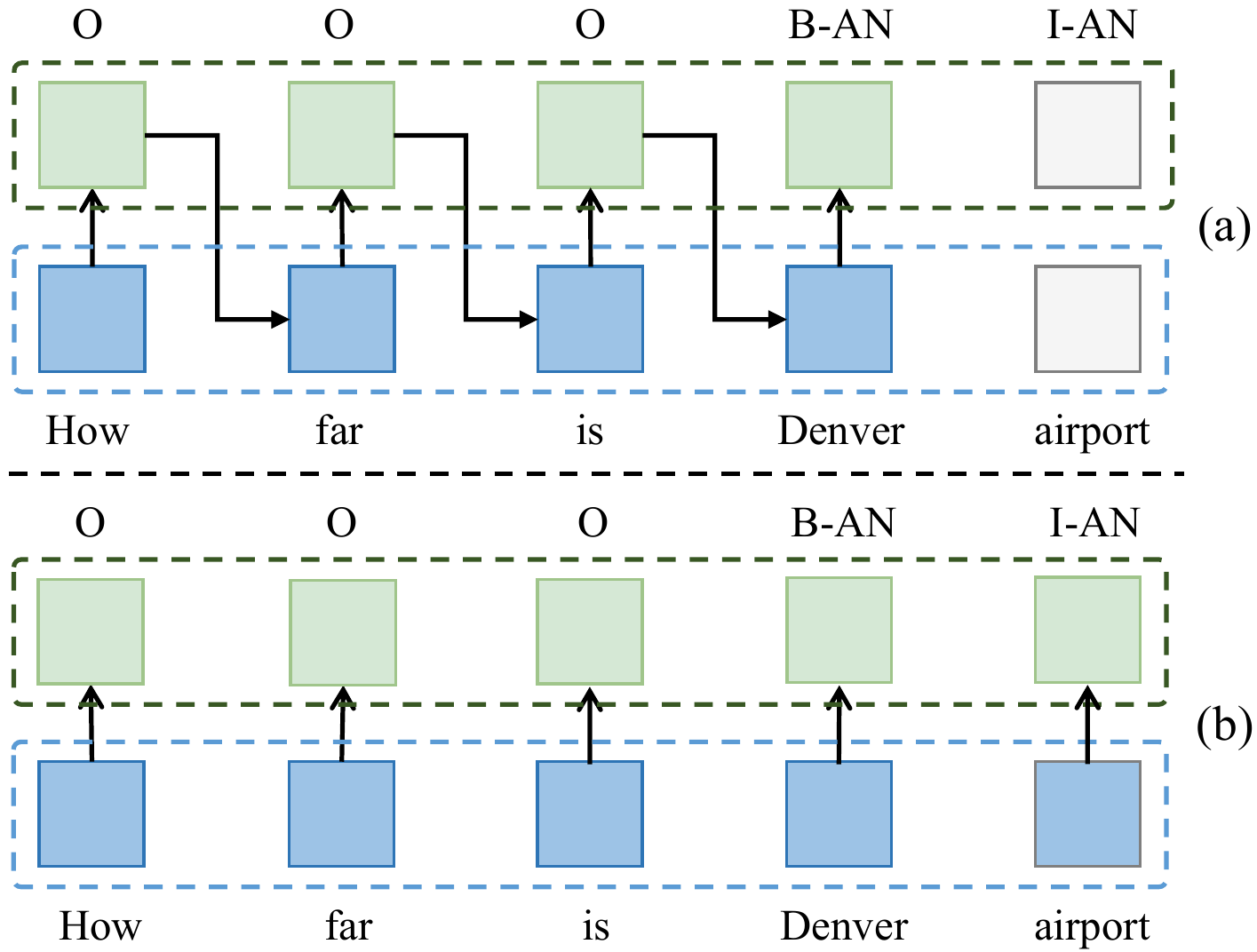}
	\caption{
		(a) Autoregressive model generates outputs word by word from left-to-right direction. The gray color denotes the unseen information when model decodes for the word Denver.  (b) Non-autoregressive model can produce outputs in parallel. \texttt{AN} denotes \texttt{airport\_name}. 
	}
	\label{fig:contrast}

\end{figure}
To this end,  ~\citet{xu2013convolutional} and ~\citet{kim2017two} begin to explore the multi-intent SLU.
However, their models only consider the multiple intent detection while ignoring slot filling task. 
Recently, \citet{gangadharaiah-narayanaswamy-2019-joint} make the first attempt to propose a multi-task framework to joint model the multiple intent detection and slot filling.
~\citet{qin-etal-2020-agif} further propose an adaptive interaction framework (AGIF) to achieve fine-grained multi-intent information integration for slot filling, obtaining state-of-the-art performance.

Though achieving the promising performance, the existing multi-intent SLU joint models heavily rely on an autoregressive fashion, as shown in Figure~\ref{fig:contrast}(a), leading to two issues:
\begin{itemize}
		\item \textit{Slow inference speed}. The autoregressive models make the generation of slot outputs must be done through the left-to-right pass, which cannot achieve parallelizable, leading to slow inference speed.
	\item \textit{Information leakage}. Autoregressive models predict each word slot conditioned on the previously generated slot information (from left-to-right), resulting in leaking the bidirectional context information.
\end{itemize}

In this paper, we explore a non-autoregressive framework for joint multiple intent detection and slot filling, with the goal of accelerating inference speed while achieving high accuracy, which is shown in Figure~\ref{fig:contrast}(b).
To this end, we propose a \textbf{G}lobal-\textbf{L}ocally \textbf{G}raph-\textbf{I}nteraction \textbf{N}etwork (GL-GIN) where the core module is a proposed local slot-aware graph layer and global intent-slot interaction layer, which achieves to generate intents and slots sequence simultaneously and non-autoregressively.
In GL-GIN, a local slot-aware graph interaction layer where each slot hidden states connect with each other is proposed to explicitly model slot dependency, in order to alleviate uncoordinated slot problem (e.g., \textit{B-singer followed by I-song})~\citep{wu-etal-2020-slotrefine} due to the non-autoregressive fashion.
A global intent-slot graph interaction layer is further introduced to perform sentence-level intent-slot interaction.
Unlike the prior works that only consider the token-level intent-slot interaction, the global graph is constructed of all tokens with multiple intents, achieving to generate slots sequence in parallel and speed up the decoding process.

Experimental results on two public datasets MixSNIPS~\citep{coucke2018snips} and MixATIS~\citep{hemphill-etal-1990-atis} show that our framework not only obtains state-of-the-art performance but also enables decoding in parallel.
In addition, we explore the pre-trained model (i.e., Roberta~\citep{DBLP:journals/corr/abs-1907-11692}) in our framework.

In summary, the contributions of this work can be concluded as follows: (1) To the best of our knowledge, we make the first attempt to explore a non-autoregressive approach for joint multiple intent detection and slot filling; (2) We propose a global-locally graph-interaction network, where the local graph is used to handle uncoordinated slots problem while a global graph is introduced to model sequence-level intent-slot interaction; (3)    Experiment results on two benchmarks show that our framework not only achieves the state-of-the-art performance but also considerably speeds up the slot decoding (up to $\times11.5$); (4) Finally, we explore the pre-trained model in our framework.
	With the pre-trained model, our model reaches a new state-of-the-art level.

For reproducibility, our code for this paper is publicly available at \url{https://github.com/yizhen20133868/GL-GIN}.

\section{Problem Definition}
\label{problem definition}
\paragraph{Multiple Intent Detection}
 Given input sequence $x$ = ($x_{1}, \dots, x_{n}$), multiple intent detection can be defined as a multi-label classification task that outputs a sequence intent label $o^{I}$ = ($o_{1}^{I}, \dots, o_{m}^{I}$), where $m$ is the number of intents in given utterance and $n$ is the length of utterance.

\paragraph{Slot Filling}
Slot filling can be seen as a sequence labeling task that maps the input utterance $x$ into a slot output sequence $o^{S}$ = ($o_{1}^{S}, \dots, o_{n}^{S}$).

\section{Approach}
\label{Approach}
\begin{figure*}[t]
	\centering
	\includegraphics[width=0.95\textwidth]{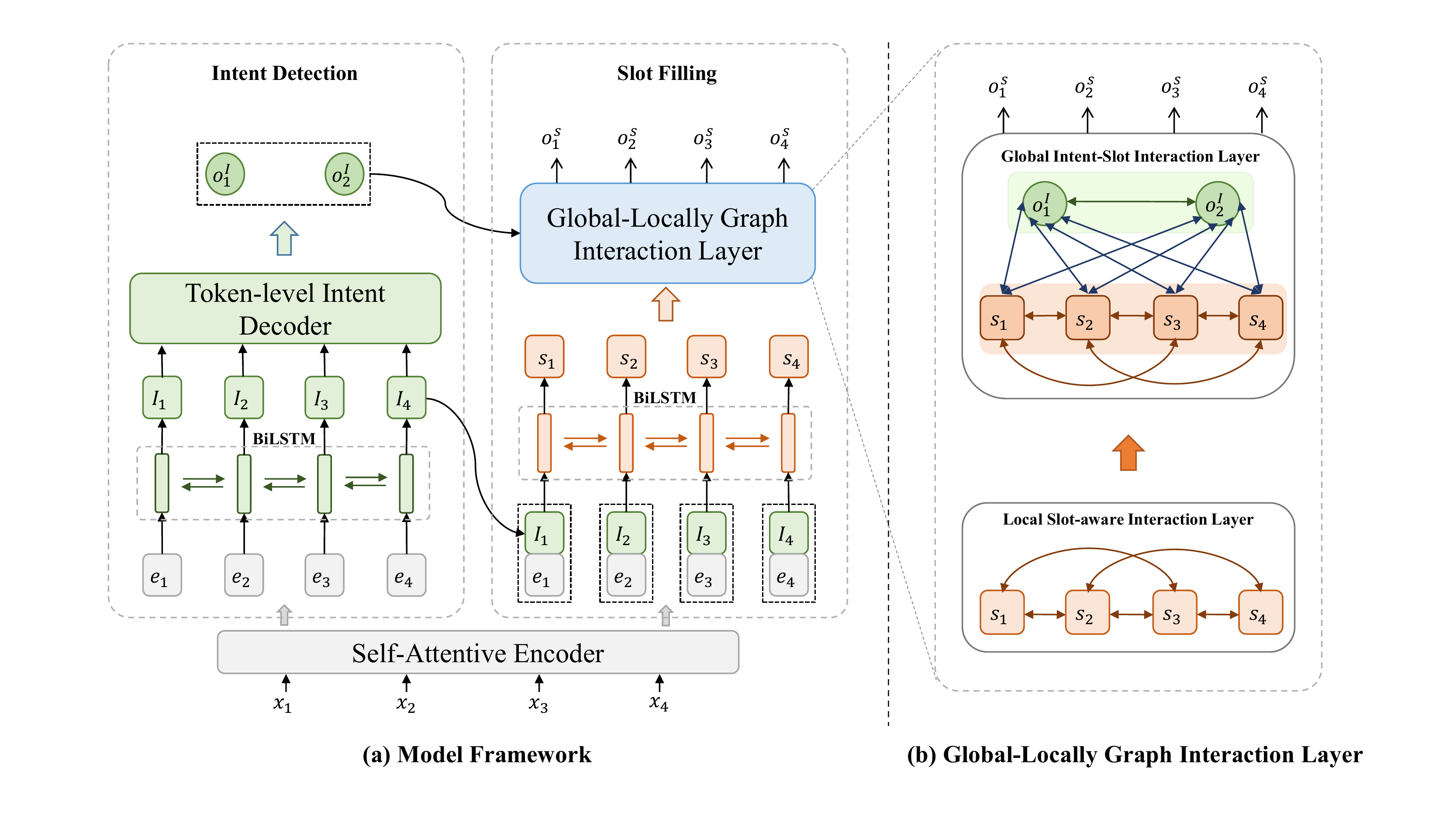}
	\caption{
	The overflow of model architecture (a) and global-locally graph interaction layer (b).
	}
	\label{fig:main}
\end{figure*}
As shown in Figure~\ref{fig:main}(a), we describe the proposed framework, which consists of a shared self-attentive encoder ($\S\ref{model:self_attentive}$), a token-level intent detection decoder ($\S\ref{model:intent_decoder}$) and a global-local graph-interaction graph decoder for slot filling ($\S\ref{model:slot_}$).
Both intent detection and slot filling are optimized simultaneously via a joint learning scheme.

\subsection{Self-attentive Encoder}\label{model:self_attentive}
Following \citet{qin-etal-2019-stack}, we utilize a self-attentive encoder with BiLSTM and self-attention mechanism to obtain the shared utterance representation, which can incorporate temporal features within word orders and contextual information.

\paragraph{BiLSTM}
The bidirectional LSTM (BiLSTM)~\citep{hochreiter1997long} have been successfully applied to sequence labeling tasks ~\citep{li-etal-2020-handling,li2021empirical}. We adopt BiLSTM to read the input sequence $\{{{x}}_{1}, {{x}}_{2}, \ldots, {{x}}_{n}\}$ forwardly and backwardly to produce context-sensitive hidden states $\boldsymbol{H} = \{\boldsymbol{h}_{1}, \boldsymbol{h}_2, \ldots, \boldsymbol{h}_{n}\}$, by repeatedly applying the $\boldsymbol{h_{i}}$ = BiLSTM ($\phi^{emb}(x_{i})$, $\boldsymbol{h}_{i-1}$, $\boldsymbol{h}_{i+1}$), where $\phi^{emb}$ is embedding function.

\paragraph{Self-Attention} 
Following \citet{NIPS2017_3f5ee243}, we map the matrix of input vectors $\boldsymbol{X} \in \mathbb R^{n \times d}$ ($d$ represents the mapped dimension) to queries $\boldsymbol{Q}$, keys $\boldsymbol{K}$ and values $\boldsymbol{V}$ matrices by using different linear projections. Then, the self-attention output $\boldsymbol{C} \in \mathbb R^{n \times d}$ is a weighted sum of values:

\begin{equation}
\boldsymbol{C} 
= \operatorname{softmax}\left( \frac{\boldsymbol{Q}\boldsymbol{K}^{\top}}{\sqrt{d_k}}\right){\boldsymbol{V}}.
\end{equation}

We concatenate the output of BiLSTM and self-attention as the final encoding representation:
\begin{equation}
\boldsymbol{ E} =\boldsymbol{ H } \mathop{||} \boldsymbol{ C},
\end{equation}
where $\boldsymbol{E}$ = $\{\boldsymbol{e}_{1},\ldots,\boldsymbol{e}_{n}\} \in \mathbb{R}^{n\times 2d}$ and $\mathop{||} $ is concatenation operation.

\subsection{Token-Level Intent Detection Decoder}\label{model:intent_decoder}
\label{model:intent}
Inspired by~\citet{qin-etal-2019-stack}, we perform a token-level multi-label multi-intent detection, where we predict multiple intents on each token and the sentence results are obtained by voting for all tokens.
Specifically, we first feed the contextual encoding $\boldsymbol{ E}$ into an intent-aware BiLSTM to enhance its task-specific representations:
\begin{eqnarray}
{\boldsymbol{h}}_ { t }^{I} = \operatorname{BiLSTM} \left( {\boldsymbol{e}} _ { t }, {\boldsymbol{h}} _ { t - 1 }^{I} , {\boldsymbol{h}} _ { t + 1 }^{I}    \right).
\end{eqnarray}
Then, $ {\boldsymbol{h}}_ { t }^{I} $ is used for intent detection, using:
\begin{eqnarray}
{I}_t \!=\! \sigma(\boldsymbol{W}_{I}(\operatorname{LeakyReLU}(\boldsymbol{W}_{h}~\boldsymbol{h}_t^{I} \!+\! \boldsymbol{b}_h))\!+\! \boldsymbol{b}_I),
\end{eqnarray}
where $I_{t}$ denotes the intent results at the $t$-th word; $\sigma$ denotes the sigmoid activation function; $\boldsymbol{ W}_{h}$ and $\boldsymbol{ W}_{I}$ are the trainable matrix parameters. 

Finally, the sentence intent results ${o}_{k}^{I}$ can be obtained by:
\begin{equation}
o^{I} = \{{o}_{k}^{I} | (\sum_{i=1}^{n}{\mathds{1}[I_{\left(i, k\right)} > 0.5]}) >   n/2 \} ,
\end{equation}
where $I_{(i, k)}$ represents the classification result of token $i$ for ${o}_{k}^{I}$. 

We predict the label as the utterance intent when it gets more than half positive predictions in all $n$ tokens.
For example, if $I_{1} = \{0.9, 0.8, 0.7, 0.1\}$, $I_{2} = \{0.8, 0.2, 0.7, 0.4\}$, $I_{3} = \{0.9, 0.3, 0.2, 0.3\}$, 
from three tokens, we get $\{3,2,1,0\}$ positive votes ($>0.5$) for four intents respectively.
Thus the index where more than half of the votes ( $>3/2$ ) were obtained was ${o}_{1}^{I}$ and ${o}_{3}^{I}$, we predict intents $o^{I} = \{{o}_{1}^{I}, {o}_{3}^{I}\}$.

\subsection{Slot Filling Decoder}\label{model:slot_}
One main advantage of our framework is the proposed global-locally graph interaction network for slot filling, which is a non-autoregressive paradigm, achieving the slot filling decoding in parallel.
In the following, we first describe the slot-aware LSTM ($\S\ref{model:slot-aware-lstm}$) to obtain the slot-aware representations, and then show how to apply the global-locally graph interaction layer ($\S\ref{model:global-locally}$) for decoding.
\subsubsection{Slot-aware LSTM}\label{model:slot-aware-lstm}
We utilize a BiLSTM to produce the slot-aware hidden representation $\boldsymbol{S}$ = ($\boldsymbol{s}_{1}$, \dots, $\boldsymbol{s}_{n}$).
At each decoding step $t$, the decoder state ${\boldsymbol{s}}_{t}$ calculating by:
\begin{equation}
{\boldsymbol{s}}_{t} = \operatorname{BiLSTM} \left({{\boldsymbol{I}}_{t} \mathop{||} {\boldsymbol{e}}_{t}, \boldsymbol{s}}_{t-1}, {\boldsymbol{s}}_{t+1} \right),
\end{equation}

where ${\boldsymbol{e}}_{t}$ denotes the aligned encoder hidden state and $\boldsymbol{I}_{t} $ denotes the predicted intent information.

\subsubsection{Global-locally Graph Interaction Layer}\label{model:global-locally}
The proposed global-locally graph interaction layer consists of two main components: one is a local slot-aware graph interaction network to model dependency across slots and another is the proposed global intent-slot graph interaction network to consider the interaction between intents and slots.

In this section, we first describe the vanilla graph attention network. Then, we illustrate the local slot-aware and global intent-slot graph interaction network, respectively.

\paragraph{Vanilla Graph Attention Network}
A graph attention network (GAT)~\citep{velivckovic2017graph} is a variant of graph neural network, which fuses the graph-structured information and
node features within the model.
 Its masked self-attention layers allow a node to attend to neighborhood features and learn different attention weights, which can automatically determine the importance and relevance between the current node with its neighborhood.

In particular, for a given graph with $N$ nodes, one-layer GAT take the initial node features $\tilde{\boldsymbol{H}} = \{\tilde{\boldsymbol{h}}_{1},\ldots, \tilde{\boldsymbol{h}}_{N}\}, \tilde{\boldsymbol{h}}_{n} \in \mathbb{R}^{F}$ as input, aiming at producing more abstract representation, $\tilde{\boldsymbol{H}}^\prime = \{\tilde{\boldsymbol{h}}^{\prime}_{1},\ldots,\tilde{\boldsymbol{h}}^{\prime}_{N}\}, \tilde{\boldsymbol{h}}^{\prime}_{n} \in \mathbb{R}^{F^\prime}$, as its output.  The attention mechanism of a typical GAT
can be summarized as below:
\begin{eqnarray}
		\tilde{\boldsymbol{h}}^{\prime}_i &= \mathop{||}_{k=1}^{K} \sigma\big(\sum_{j \in \mathcal{N}_i} \alpha_{ij}^k \boldsymbol{W}_h^k \tilde{\boldsymbol{h}}_{j}\big) , \\
			\alpha_{ij} &= \frac{\exp(\operatorname{LeakyReLU}\left(\mathbf{a}^\top[\boldsymbol{W}_h\tilde{\boldsymbol{h}}_i\|\boldsymbol{W}_h\tilde{\boldsymbol{h}}_j]\right))}{\sum_{j^\prime \in \mathcal{N}_i} \exp{(\operatorname{LeakyReLU}\left(\mathbf{a}^\top[\boldsymbol{W}_h\tilde{\boldsymbol{h}}_i\|\boldsymbol{W}_h\tilde{\boldsymbol{h}}_j^\prime]\right))}} , 
\end{eqnarray}
where $\boldsymbol{W}_h \in \mathbb{R}^{F^\prime \times F}$ and $\mathbf{a} \in \mathbb{R}^{2F^\prime}$ are the trainable weight matrix; $\mathcal{N}_i$ denotes the neighbors of node $i$ (including $i$); $\alpha_{ij}$ is the normalized attention coefficients and $\sigma$ represents the nonlinearity activation function; $K$ is the number of heads. 

\paragraph{Local Slot-aware Graph Interaction Layer}
Given slot decode hidden representations $\boldsymbol{S}$ = ($\boldsymbol{s}_{1}$, \dots, $\boldsymbol{s}_{n}$), we construct a local slot-aware graph where each slot hidden node connects to other slots.
This allows the model to achieve to model the dependency across slots, alleviating the uncoordinated slots problem.
Specifically, we construct the graph $\mathcal{G}=(V,\mathcal{E})$  in the following way,

\subparagraph{Vertices}We define the $V$ as the vertices set. Each word slot is represented as a vertex.
Each vertex is initialized with the corresponding slot hidden representation.
Thus, the first layer states vector for all nodes is $\boldsymbol{S}^{1}$ = $\boldsymbol{S}$ = ($\boldsymbol{s}_{1}$, \dots, $\boldsymbol{s}_{n}$).

\subparagraph{Edges}
Since we aim to model dependency across slots, we construct a slot-aware graph interaction layer so that the dependency relationship can be propagated from neighbor nodes to the current node. Each slot can connect other slots with a window size. For node $\boldsymbol{S}_{i}$, only $\{\boldsymbol{S}_{i-{m}},\ldots,\boldsymbol{S}_{i+{m}}\} $ will be connected where $m$ is a hyper-parameter denotes the size of sliding window that controls the length of utilizing utterance context.
\subparagraph{Information Aggregation}The aggregation process at $l$-th layer can be defined as:
\begin{equation}
\boldsymbol{s}^{l+1}_i =   \sigma\big(\sum_{j  \in \mathcal{N}_i} \alpha_{ij} \boldsymbol{W}_l \boldsymbol{s}^{l}_j\big) ,
\end{equation}
where $\mathcal{N}_i$ is a set of vertices that denotes the connected slots.

After stacking $L$ layer, we obtain the contextual slot-aware local hidden features $\boldsymbol{S}^{L+1}$ =$\{\boldsymbol{s}^{L+1}_{1},\ldots,\boldsymbol{s}^{L+1}_{n}\} $

\paragraph{Global Slot-Intent Graph Interaction Layer}
To achieve sentence-level intent-slot interaction, we construct a global slot-intent interaction graph where all predicted multiple intents and sequence slots are connected, achieving to output slot sequences in parallel.
Specifically, we construct the graph $\mathcal{G}=(V,\mathcal{E})$  in the following way,
\subparagraph{Vertices}
As we model the interaction between intent and slot token, we have $n+m$ number of nodes in the graph where $n$ is the sequence length and $m$ is the number of intent labels predicted by the intent decoder. The input of slot token feature is ${\boldsymbol{G}^{[S, 1]}}$ = $\boldsymbol{S}^{L+1}$ =$\{\boldsymbol{s}^{L+1}_{1},\ldots,\boldsymbol{s}^{L+1}_{n}\} $ which is produced by slot-aware local interaction graph network while the input intent feature is an embedding ${\boldsymbol{G}^{[I, 1]}}$ =  $\{\phi^{emb}(\boldsymbol{o}_{1}^{I}), \ldots, \phi^{emb}(\boldsymbol{o}_{m}^{I})\}$ where $\phi^{emb}$ is a trainable embedding matrix. The first layer states vector for slot and intent nodes is $\boldsymbol{G}^{1}$ = \{${\boldsymbol{G}^{[I, 1]}}$ , ${\boldsymbol{G}^{[S, 1]}}$ \} = \{$\phi^{emb}(\boldsymbol{o}_{1}^{I}), \ldots, \phi^{emb}(\boldsymbol{o}_{m}^{I})$, $\boldsymbol{s}^{L+1}_{1},\ldots,\boldsymbol{s}^{L+1}_{n} $\}
\subparagraph{Edges}
There are three types of connections in this graph network.
\begin{itemize}
	\item \textit{intent-slot connection}: 
	Since slots and intents are highly tied, we construct the intent-slot connection to model the interaction between the two tasks.
	Specifically, each slot connects all predicted multiple intents to automatically capture relevant intent information.
		\item  \textit{slot-slot connection}: 
	We construct the slot-slot connection where each slot node connects other slots with the window size to further model the slot dependency and incorporate the bidirectional contextual information.
	\item \textit{intent-intent connection}: 
	Following~\citet{qin-etal-2020-agif}, we connect all the intent nodes to each other to model the relationship between each intent, since all of them express the same utterance’s intent.
\end{itemize}

\subparagraph{Information Aggregation} The aggregation process of the global GAT layer can be formulated as:

\begin{equation}
\boldsymbol{g}^{[S, l+1]}_i = \sigma(\sum_{j \in \boldsymbol{\mathcal{G}}^{S}} \alpha_{ij} \boldsymbol{W}_g \boldsymbol{g}^{[S, l]}_j + \sum_{j \in \boldsymbol{\mathcal{G}}^I} \alpha_{ij} \boldsymbol{W}_g \boldsymbol{g}^{[I,l]}_j),
\end{equation}
where $\mathcal{G}^{S}$ and $\mathcal{G}^{I}$ are vertices sets which denotes the connected slots and intents, respectively.

\subsubsection{Slot Prediction}
After $L$ layers' propagation, we obtain the final slot representation $\boldsymbol{G}^{[S,L+1]}$ for slot prediction.
\begin{align}
{\boldsymbol{y}}_{t}^{S} &= \operatorname{softmax} \left({\boldsymbol{W}}_{s}{\boldsymbol{g}}^{[S,L+1]}_{t}\right),\\
\boldsymbol{o}_{t}^{S} &= \arg \max({\boldsymbol{y}}_{t}^{S}),
\end{align}
where $\boldsymbol{W}_{s}$ is a trainable parameter and $\boldsymbol{o}_{t}^{S}$ is the predicted slot if the \textit{t}-th token in an utterance.

\subsection{Joint Training}
Following \citet{goo-etal-2018-slot}, we adopt a joint training model to consider the two tasks and update parameters by joint optimizing.
The intent detection objective is:
\begin{eqnarray}
\operatorname{CE}(\hat{y}, y) = \hat{y} \log \left({y}\right)+\left(1-\hat{y}\right) \log \left(1-{y}\right), \\
\mathcal{L}_{1} \triangleq -\sum_{i=1}^{n}\sum_{j=1}^{N_I} \operatorname{CE}(\hat{y}_{i}^{(j,I)}, {y}_{i}^{(j,I)}) \,.
\end{eqnarray}

Similarly, the slot filling task objective is:
\begin{equation}
\mathcal{L}_{2} \triangleq - \sum_{i=1}^{n}\sum_{j=1}^{N_S}{{\hat{{y}}_{i}^{(j,S)}}\log \left({{y}}_{i}^{(j,S)}\right)},
\end{equation}
where $N_I$ is the number of single intent labels and $N_S$ is the number of slot labels.

The final joint objective is formulated as:
\begin{equation}
\mathcal{L}=\alpha \mathcal{L}_{1} +\beta  \mathcal{L}_{2},
\end{equation}
where $\alpha$ and $\beta$ are hyper-parameters.

\section{Experiments}
\label{experiments}
\begin{table*}[h]
	\centering
	\begin{adjustbox}{width=\textwidth}
		\begin{tabular}{l||ccc||ccc}
			\hline
			\multirow{2}{*}{\textbf{Model}} & \multicolumn{3}{c||}{\textbf{MixATIS}} & \multicolumn{3}{c}{\textbf{MixSNIPS}} \\
			\cline{2-7}
			& Overall(Acc) & Slot(F1) & Intent(Acc) & Overall(Acc) & Slot(F1)& Intent(Acc) 
			\\
			\hline
			{Attention BiRNN \citep{Liu+2016}} & 39.1& 86.4 & 74.6  & 59.5 & 89.4  & 95.4  \\
			{Slot-Gated  \citep{goo-etal-2018-slot}} & 35.5& 87.7  & 63.9  & 55.4& 87.9& 94.6  \\
			{Bi-Model \citep{wang-etal-2018-bi}}  & 34.4& 83.9  & 70.3 & 63.4& 90.7 & 95.6  \\
			{SF-ID \citep{e-etal-2019-novel}} & 34.9& 87.4 & 66.2 & 59.9 & 90.6  & 95.0  \\
			{Stack-Propagation \citep{qin-etal-2019-stack}}  & 40.1 & 87.8  & 72.1& 72.9& 94.2& 96.0  \\
			{Joint Multiple ID-SF \citep{gangadharaiah-narayanaswamy-2019-joint}} & 36.1 & 84.6  & 73.4 & 62.9& 90.6  & 95.1  \\
			{AGIF \citep{qin-etal-2020-agif}}& 40.8& 86.7  & 74.4  & 74.2& 94.2  & 95.1  \\
			\hline
			{GL-GIN} & \textbf{43.5*}& \textbf{88.3*} & \textbf{76.3*} & \textbf{75.4*} & \textbf{94.9*}  & {95.6} 
			\\ 
			\hline 
		\end{tabular}
	\end{adjustbox}
	\caption{Main results. 
				The numbers with * indicate that the improvement of our framework over all baselines is statistically significant with $p<0.05$ under t-test.
	} 
	\label{tab:main results}
\end{table*}

\subsection{Datasets}
We conduct experiments on two publicly available multi-intent datasets.\footnote{We adopt the cleaned verison that removes the repeated sentences in original dataset, which is available at \url{https://github.com/LooperXX/AGIF}.} One is the MixATIS~\citep{hemphill-etal-1990-atis, qin-etal-2020-agif}, which includes 13,162 utterances for training,  756 utterances for validation and 828 utterances for testing.
Another is MixSNIPS~\citep{coucke2018snips, qin-etal-2020-agif}, with 39,776, 2,198, 2,199 utterances for training, validation and testing.

\subsection{Experimental Settings}
The dimensionality of the embedding is 128 and 64 on ATIS and SNIPS, respectively. The dimensionality of the LSTM hidden units is 256.
The batch size is 16.
The number of the multi head is 4 and 8 on MixATIS and MixSNIPS dataset, respectively. 
All layer number of graph attention network is set to 2. We use Adam~\citep{DBLP:journals/corr/KingmaB14} to optimize the parameters in our model. 
For all the experiments, we select the model which works the best on the dev set and then evaluate it on the test set.
All experiments are conducted at GeForce RTX 2080Ti and TITAN Xp. 
\subsection{Baselines}
We compare our model with the following best baselines:
(1) {\texttt{Attention BiRNN.}} \citet{Liu+2016} propose an alignment-based RNN for joint slot filling and intent detection;
(2)  {\texttt{Slot-Gated Atten.}} \citet{goo-etal-2018-slot} propose a slot-gated joint model, explicitly considering the correlation between slot filling and intent detection;
(3) {\texttt{Bi-Model.}} \citet{wang-etal-2018-bi} propose the Bi-model to model the bi-directional between the intent detection and slot filling; 
(4) {\texttt{SF-ID Network.}} \citet{e-etal-2019-novel} proposes the SF-ID network to establish a direct connection between the two tasks;
(5) {\texttt{Stack-Propagation.}}  \citet{qin-etal-2019-stack} adopt a stack-propagation framework to explicitly incorporate intent detection for guiding slot filling;
(6) {\texttt{Joint Multiple ID-SF}}.  \citet{gangadharaiah-narayanaswamy-2019-joint} propose
a multi-task framework with slot-gated mechanism for multiple intent detection and slot filling;
(7) {\texttt{AGIF}}  \citet{qin-etal-2020-agif} proposes an adaptive interaction network to achieve the fine-grained multi-intent information integration, achieving state-of-the-art performance.
 
\begin{table}[t]
	\centering
	\begin{adjustbox}{width=0.45\textwidth}
		\begin{tabular}{l|cccl}
			\hline
			Model & \multicolumn{1}{l}{Decode Latency(s)} & \multicolumn{1}{l}{Speedup}\\ 
			\hline
			Stack-Propagation      & 34.5 &     8.2$\times$             &     \\
			Joint Multiple ID-SF & 45.3 &  10.8$\times$             &  \\
			AGIF                    & 48.5 &     11.5$\times$              &     \\
			\hline
			GL-GIN               & 4.2 &      1.0$\times$    &               \\
			\hline
		\end{tabular}
	\end{adjustbox}
	\caption{Speed comparison. Speedup is based on the ratio of the time taken by the slot decoding part of different models to run an epoch on the MixATIS dataset with batch size set to 32.} 
	\label{tab:speed results}
\end{table}

\subsection{Main Results}

\begin{table*}[h] 
	\small
	\centering
	\begin{adjustbox}{width=\textwidth}
		\begin{tabular}{l||ccc||ccc}
			\hline
			\multirow{2}{*}{\textbf{Model}} & \multicolumn{3}{c||}{\textbf{MixATIS}} & \multicolumn{3}{c}{\textbf{MixSNIPS}} \\
			\cline{2-7}
			& Overall(Acc)& Slot(F1) & Intent(Acc) & Overall(Acc) & Slot(F1)  & Intent(Acc) 
			\\
			\hline
			{w/o Local Slot-Aware  GAL} & 41.1& 86.8  & 74.0  & 71.4 & 93.7  & 95.2 \\
			{w/o Global Intent-Slot GAL}& 40.9 & 87.4 & 75.5 & 71.7 & 93.6  & 95.5  \\
				\quad+ More Parameters & 41.9 & 87.7  & 75.0& 73.0  & 93.8  & 95.5  \\
			{w/o Global-locally GAL} & 40.5 & 86.3  & 75.2 & 70.2 & 92.9  & 95.0 \\
			\hdashline
			{GL-GIN} & 43.5 & 88.3  & 76.3  & 75.4& 94.9 & 95.6
			\\ 
			\hline
		\end{tabular}
	\end{adjustbox}
	\caption{Ablation Experiment.
	} 
	\label{tab:support}
\end{table*}

Following \citet{goo-etal-2018-slot} and \citet{qin-etal-2020-agif}, we evaluate the performance of slot filling using F1 score, intent prediction using accuracy,
the sentence-level semantic frame parsing using overall accuracy.
Overall accuracy measures the ratio of sentences for which both intent and slot are predicted correctly in a sentence. 

Table~\ref{tab:main results} shows the results, we have the following observations: (1) On slot filling task, our framework outperforms the best baseline \texttt{AGIF} in F1 scores on two datasets, which indicates the proposed local slot-aware graph successfully models the dependency across slots, so that the slot filling performance can be improved.
(2) More importantly, compared with the \texttt{AGIF}, our framework achieves +2.7\% and 1.2\% improvements for MixATIS and MixSNIPS on overall accuracy, respectively.
We attribute it to the fact that our proposed global intent-slot interaction graph can better capture the correlation between intents and slots, improving the SLU performance.

\subsection{Analysis}
\subsubsection{Speedup}
One of the core contributions of our framework is that the decoding process of slot filling can be significantly accelerated with the proposed non-autoregressive mechanism.
We evaluate the speed by running the model on the MixATIS test data in an epoch, fixing the batch size to 32.
The comparison results are shown in Table~\ref{tab:speed results}.
We observe that our model achieves the $\times$8.2, $\times$10.8 and $\times$11.5 speedup compared with  
SOTA models \texttt{stack-propagation}, \texttt{Joint Multiple ID-SF} and \texttt{AGIF}.
This is because that their model utilizes an autoregressive architecture that only performs slot filling word by word, while our non-autoregressive framework can conduct slot filling decoding in parallel.
In addition, it's worth noting that as the batch size gets larger, \texttt{GL-GIN} can achieve better acceleration where our model could achieve $\times$17.2 speedup compared with \texttt{AGIF} when batch size is 64.

\subsubsection{Effectiveness of the Local Slot-aware Graph Interaction Layer}
We study the effectiveness of the local slot-aware interaction graph layer with the following ablation. We remove the local graph interaction layer and directly feed the output of the \textit{slot LSTM} to the global intent-slot graph interaction layer. We refer it to \textit{w/o local GAL} in Tabel~\ref{tab:support}. We can clearly observe that the slot F1 drops by 1.5\% and 1.2\% on MixATIS and MixSNIPS datasets. We attribute this to the fact that local slot-aware GAL can capture the slot dependency for each token, which helps to alleviate the slot uncoordinated problems.
A qualitative analysis can be founded at Section~\ref{exp:case}.

\begin{figure}[t]
	\centering
	\includegraphics[width=0.3\textwidth]{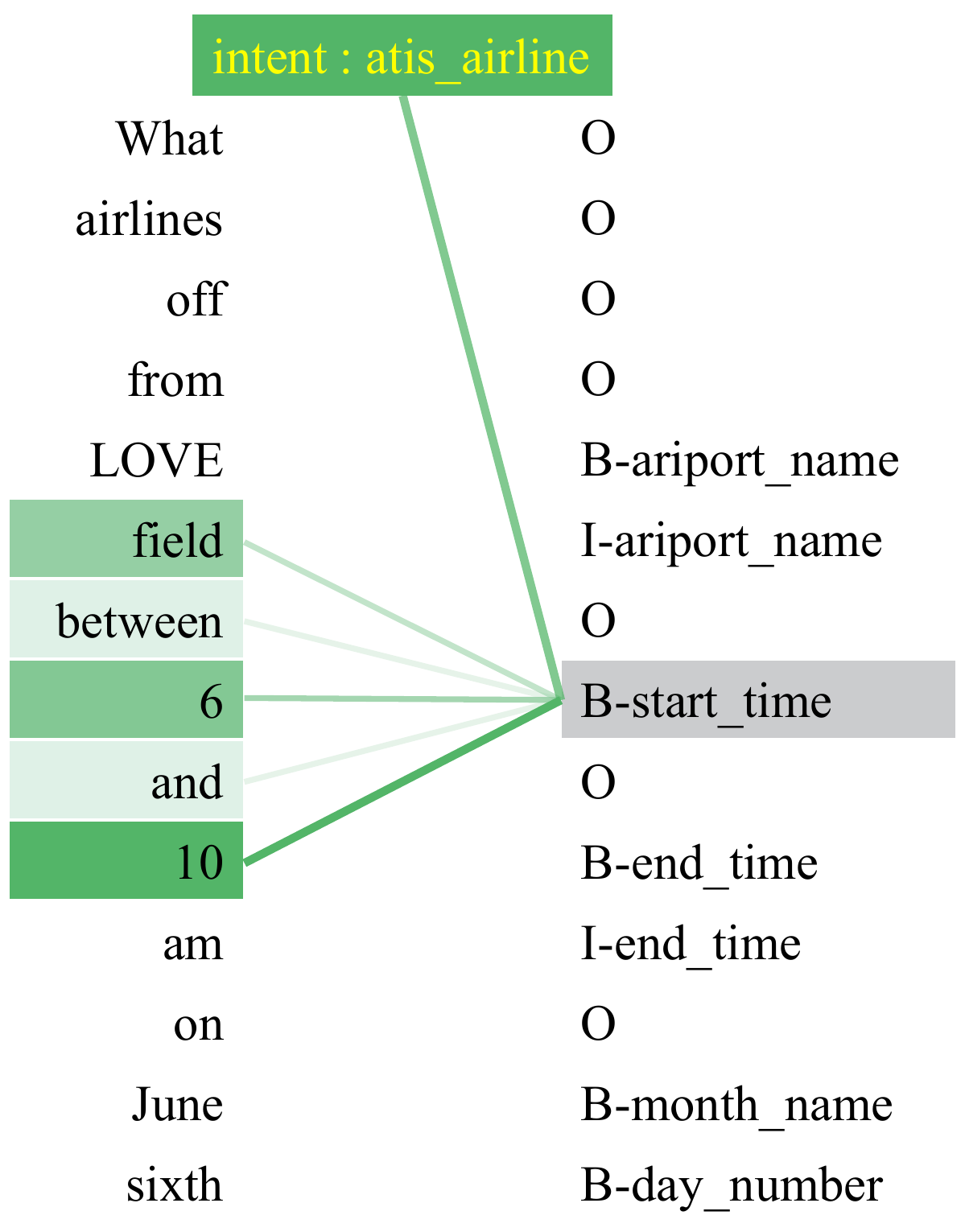}
	\caption{
	Visualization. We use the green color to indicate the attention value. 
	}
	\label{fig:visualize}

\end{figure}

\subsubsection{Effectiveness of Global Slot-Intent Graph Interaction Layer}
In order to verify the effectiveness of slot-intent global interaction graph layer, we remove the global interaction layer and utilizes the output of local slot-aware GAL module for slot filling. It is named as \textit{w/o Global Intent-slot GAL} in Table~\ref{tab:support}.  We can observe that  the slot f1 drops by 0.9\%, 1.3\%, which demonstrates that intent-slot graph interaction layer can capture the correlation between multiple intents, which is beneficial for the semantic performance of SLU system.

Following~\citet{qin-etal-2020-agif}, we replace multiple LSTM layers (2-layers) as the proposed global-locally graph layer to verify that the proposed global-locally graph interaction layer rather than the added parameters works.
 Table~\ref{tab:support} (\textit{more parameters}) shows the results.
 We observe that our model outperforms \textit{more parameters} by 1.6\% and 2.4\% overall accuracy in two datasets, which shows that the improvements come from the proposed Global-locally graph interaction layer rather than the involved parameters.

\begin{table*}[h]
	\centering
	\begin{adjustbox}{width=\textwidth}
		\begin{tabular}{l||cccccccccccccc}
			\hline
			\textbf{texts} & What & airlines & off & from & LOVE & field & between & 6 & and & 10 & am & on & June & sixth \\
			\hline
			\hline
			\textbf{AGIF} & O & O & O & O & 
			\begin{tabular}[c]{@{}l@{}}B-fromloc\\airport\_name\end{tabular} & \begin{tabular}[c]{@{}l@{}}I-fromloc\\airport\_name\end{tabular} & 
			O & 
			\textcolor[rgb]{1,0,0}{O} & O & 
			\begin{tabular}[c]{@{}l@{}}B-depart\_time\\end\_time\end{tabular} & 
			\textcolor[rgb]{1,0,0}{\begin{tabular}[c]{@{}l@{}}I-toloc\\airport\_name\end{tabular}} & 
			O & 
			
			\begin{tabular}[c]{@{}l@{}}B-depart\_date\\month\_name\end{tabular} & 
			\begin{tabular}[c]{@{}l@{}}B-depart\_date\\day\_number\end{tabular}\\
			\hline

			\textbf{GL-GIN} & O & O & O & O & 
			\begin{tabular}[c]{@{}l@{}}B-fromloc\\airport\_name\end{tabular} & \begin{tabular}[c]{@{}l@{}}I-fromloc\\airport\_name\end{tabular} & 
			O & 
			\begin{tabular}[c]{@{}l@{}}B-depart\_time\\start\_time\end{tabular} & O & 
			\begin{tabular}[c]{@{}l@{}}B-depart\_time\\end\_time\end{tabular} & 
			\begin{tabular}[c]{@{}l@{}}I-depart\_time\\end\_time\end{tabular} & 
			O & 
			
			\begin{tabular}[c]{@{}l@{}}B-depart\_date\\month\_name\end{tabular} & 
			\begin{tabular}[c]{@{}l@{}}B-depart\_date\\day\_number\end{tabular}\\
			\hline

			\hline
		\end{tabular}
	\end{adjustbox}
	\caption{Case study. Predicted slots sequence about utterance
		``What airlines off from LOVE field between 6 and 10 am on June sixth''
	} 
	\label{tab:case}
\end{table*}

\subsubsection{Effectiveness of the Global-locally Graph Interaction Layer}
 Instead of using the whole global-locally graph interaction layer for slot filling, we directly leverage the output of slot-aware LSTM to predict each token slot to verify the effect of the global-locally graph interaction layer. We name the experiment as \textit{w/o Global-locally GAL} in Tabel~\ref{tab:support}. From the results, We can observe that the absence of global GAT module leads to 3.0\% and 5.2\% overall accuracy drops on two datasets. This indicates that the global-locally graph interaction layer encourages our model to leverage slot dependency and intent information, which can improve SLU performance.
\subsubsection{Visualization}
To better understand how global-local graph interaction layer affects and contributes to the final result, we visualize the attention value of the Global intent-slot GAL.
As is shown in Figure~\ref{fig:visualize}, we visualize the dependence of the word ``\texttt{6}'' on context and intent information. We can clearly observe that token ``\texttt{6}'' obtains information from all contextual tokens. The information from ``\texttt{and 10}'' helps to predict the slot, where the prior autoregressive models cannot be achieved due to the generation word by word from left to right.
\begin{figure}[t]
	\centering
	\includegraphics[width=0.5\textwidth]{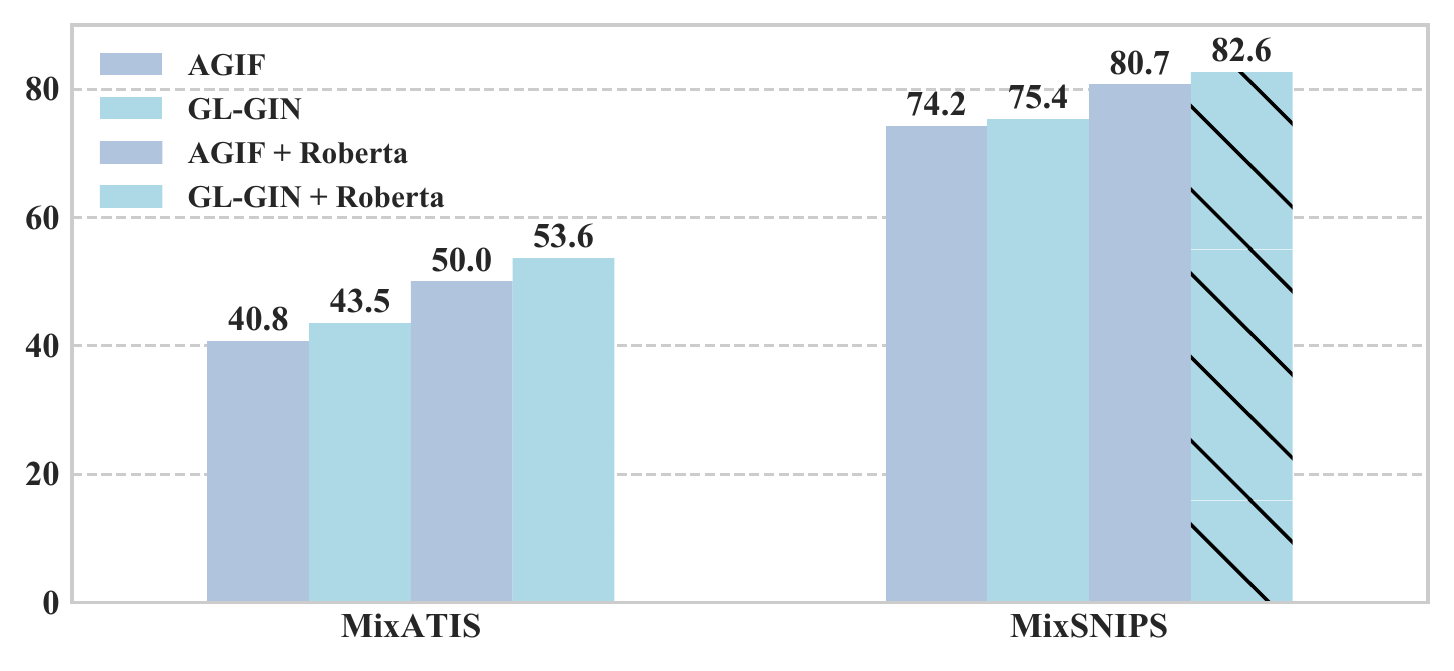}
	\caption{
		Overall accuracy Performances with \texttt{Roberta}.
	}
	\label{fig:pretrain}

\end{figure}
\subsubsection{Qualitative analysis}\label{exp:case}
We conduct qualitative analysis by providing a case study that consists of two sequence slots which are generated from \texttt{AGIF} and our model.
From Table~\ref{tab:case}, for the word ``\texttt{6}'', \texttt{AGIF} predicts its slot label as ``\texttt{O}'' incorrectly.
This is because that \texttt{AGIF} only models its left information, which makes it hard to predict ``\texttt{6}'' is a time slot.
In contrast, our model predicts the slot label correctly. We attribute this to the fact that our proposed global intent-slot interaction layer can model bidirectional contextual information.
In addition, our framework predicts the word slot ``\texttt{am}'' correctly while \texttt{AGIF} predicts it incorrectly (I-airport\_name follows B-depart\_time), indicating that the proposed local slot-aware graph layer has successfully captured the slot dependency.

\subsubsection{Effect of Pre-trained Model}

Following~\citet{qin-etal-2019-stack}, we explore the pre-trained model in our framework.
We replace the self-attentive encoder by \textit{Roberta}~\citep{DBLP:journals/corr/abs-1907-11692} with the fine-tuning approach.
We keep other components identical to our framework and follow~\citet{qin-etal-2019-stack} to consider the first subword label if a word is broken into multiple subwords.

Figure~\ref{fig:pretrain} gives the result comparison of \texttt{AGIF}, \texttt{GL-GIN} and two models with \texttt{Roberta} on two datasets.
We have two interesting observations. First, the \texttt{Roberta-based} model remarkably well on two datasets. We attribute this to the fact that pre-trained models can provide rich semantic features, which can help SLU. Second, \texttt{GL-GIN + Roberta} outperforms \texttt{AGIF+Roberta} on both datasets and reaches a new state-of-the-art performance, which further verifies the effectiveness of our proposed framework.

\section{Related Work}
\label{related work}
\paragraph{Slot Filling and Intent Detection}
Recently, joint models~\citep{10.5555/3060832.3061040,hakkani-tr2016multi-domain, goo-etal-2018-slot,li-etal-2018-self,xia-etal-2018-zero,e-etal-2019-novel,liu-etal-2019-cm,qin-etal-2019-stack, zhang-etal-2019-joint, wu-etal-2020-slotrefine,qin2021cointeractive,ni2021recent} are proposed to consider the strong correlation between intent detection and slot filling have obtained remarkable success.
Compared with their work, we focus on jointly modeling multiple intent detection and slot filling while they only consider the single-intent scenario.

More recently, multiple intent detection can handle utterances with multiple intents, which has attracted increasing attention.
To the end, ~\citet{xu2013convolutional} and ~\citet{kim2017two} begin to explore the multiple intent detection.
\citet{gangadharaiah-narayanaswamy-2019-joint} first apply a multi-task framework with a slot-gate mechanism to jointly model the multiple intent detection and slot filling.
\citet{qin-etal-2020-agif} propose an adaptive interaction network to achieve the fine-grained multiple intent information integration for token-level slot filling, achieving the state-of-the-art performance.
Their models adopt the autoregressive architecture for joint multiple intent detection and slot filling. 
In contrast, we propose a non-autoregressive approach, achieving parallel decoding.
To the best of our knowledge, we are the first to explore a non-autoregressive architecture for multiple intent detection and slot filling.
\paragraph{Graph Neural Network for NLP}
Graph neural networks that operate directly on graph structures to model the structural information, which has been applied successfully in various NLP tasks.
~\citet{linmei-etal-2019-heterogeneous} and ~\citet{huang-carley-2019-syntax} explore graph attention network (GAT) ~\citep{velivckovic2017graph} for classification task to incorporate the dependency parser information.
~\citet{cetoli-etal-2017-graph} and ~\citet{Liu_Chang_Huang_Tang_Cheung_2019} apply graph neural network to model the non-local contextual information for sequence labeling tasks.
~\citet{yasunaga-etal-2017-graph} and ~\citet{feng2020dialogue} successfully apply a graph network to model the discourse information for the summarization generation task, which achieved promising performance.
Graph structure are successfully applied for dialogue direction~\citep{feng2020incorporating,fu-etal-2020-drts,qin2020cogat,qin2021knowing}. 
In our work, we apply a global-locally graph interaction network to model the slot dependency and interaction between the multiple intents and slots.

\section{Conclusion}
\label{conclusion}
In this paper, we investigated a non-autoregressive model for joint multiple intent detection and slot filling. 
To this end, we proposed a global-locally graph interaction network where the uncoordinated-slots problem can be addressed with the proposed local slot-aware graph while the interaction between intents and slots can be modeled by the proposed global intent-slot graph.
Experimental results on two datasets show that our framework achieves state-of-the-art performance with $\times 11.5$ times faster than the prior work.

\section*{Acknowledgements}
This work was supported by the National Key R\&D Program of China via grant 2020AAA0106501 and the National Natural Science Foundation of China (NSFC) via grant 61976072 and 61772153. This work was also supported by the Zhejiang Lab’s International Talent Fund for Young Professionals.

\bibliography{anthology,acl2021}
\bibliographystyle{acl_natbib}

\appendix

\end{document}